\documentclass[twoside,11pt]{article}

\usepackage{makecell}
\usepackage{jmlr2e}
\usepackage{enumitem}
\usepackage[utf8]{inputenc}

\DeclareFixedFont{\ttb}{T1}{txtt}{bx}{n}{12} % for bold
\DeclareFixedFont{\ttm}{T1}{txtt}{m}{n}{12}  % for normal

\usepackage{color}
\definecolor{deepblue}{rgb}{0,0,0.5}
\definecolor{deepred}{rgb}{0.6,0,0}
\definecolor{deepgreen}{rgb}{0,0.5,0}

\usepackage{listings}

\newcommand\pythonstyle{\lstset{
language=Python,
basicstyle=\small,
otherkeywords={self, transe, transh},             % Add keywords here
keywordstyle=\ttb\color{deepblue},
emph={MyClass,__init__, python},          % Custom highlighting
emphstyle=\ttb\color{deepred},    % Custom highlighting style
stringstyle=\color{deepgreen},
frame=tb,                         % Any extra options here
showstringspaces=false            % 
}}

\lstnewenvironment{python}[1][]
{
\pythonstyle
\lstset{#1}
}
{}

\newcommand\pythoninline[1]{{\pythonstyle\lstinline!#1!}}

\firstpageno{1}

\begin{document}
\title{Pykg2vec: A Python Library for Knowledge Graph Embedding}

\author{\name{Shih Yuan Yu}\thanks{Shih Yuan Yu and Sujit Rokka Chhetri contributed equally to this article} \email shihyuay@uci.edu \\
        \name{Sujit Rokka Chhetri\footnotemark[1]} \email schhetri@uci.edu \\
        \addr Department of Electrical Engineering and Computer Science, University of California-Irvine
        \AND
        \name Arquimedes Canedo \email arquimedes.canedo@siemens.com \\
        \addr Siemens Corporate Technology, Princeton
        \AND
        \name Palash Goyal \email palashgo@usc.edu \\
       \addr Department of Computer Science,
       University of Southern California
        \AND
        \name{Mohammad Abdullah Al Faruque} \email alfaruqu@uci.edu\\
        \addr Department of Electrical Engineering and Computer Science, University of California-Irvine
}

\editor{}

\maketitle

\begin{abstract}%   <- trailing '%' for backward compatibility of .sty file
Pykg2vec is an open-source Python library for learning the representations of the entities and relations in knowledge graphs. Pykg2vec's flexible and modular software architecture currently implements 16 state-of-the-art knowledge graph embedding algorithms, and is designed to easily incorporate new algorithms. The goal of pykg2vec is to provide a practical and educational platform to accelerate research in knowledge graph representation learning. Pykg2vec is built on top of TensorFlow and Python's multiprocessing framework and provides modules for batch generation, Bayesian hyperparameter optimization, mean rank evaluation, embedding, and result visualization. Pykg2vec is released under the MIT License and is also available in the Python Package Index (PyPI). The source code of pykg2vec is available at \url{https://github.com/Sujit-O/pykg2vec}.

\end{abstract}


\section{Introduction}
In recent years, Knowledge Graph Embedding (KGE) methods have been applied in benchmark datasets including Wikidata (\cite{Free2014}), Freebase (\cite{Bollacker2008}), DBpedia (\cite{Auer2007}), and YAGO (\cite{Suchanek2017}). Applications of KGE methods include fact prediction, question answering, and recommender systems.

KGE is an active area of research and many authors have provided reference software implementations. However, most of these are standalone reference implementations and therefore it is difficult and time-consuming to: (i) find the source code; (ii) adapt the source code to new datasets; (iii) correctly parameterize the models; and (iv) compare against other methods. Recently, this problem has been partially addressed by libraries such as OpenKE~\cite{han2018openke} and AmpliGraph~\cite{ampligraph} that provide a framework common to several KGE methods. However, these frameworks take different perspectives, make specific assumptions, and thus the resulting implementations diverge substantially from the original architectures. Furthermore, these libraries often force the user to use preset hyperparameters, or make implicit use of \textit{golden} hyperparameters, and thus make it tedious and time-consuming to adapt the models to new datasets. 

This paper presents pykg2vec, a single Python library with 16 state-of-the-art KGE methods. The goals of pykg2vec are to be practical and educational. The practical value is achieved through: (a) proper use of GPUs and CPUs; (b) a set of tools to automate the discovery of golden hyperparameters; and (c) a set of visualization tools for the training and results of the embeddings. The educational value is achieved through: (d) a modular and flexible software architecture and KGE pipeline; and (e) access to a large number of state-of-the-art KGE models.

\section{Knowledge Graph Embedding Methods}
A knowledge graph contains a set of entities $\mathbb{E}$ and relations $\mathbb{R}$ between entities. The set of facts $\mathbb{D}^+$ in the knowledge graph are represented in the form of triples $(h, r, t)$, where $h,t\in\mathbb{E}$ are referred to as the \textit{head} (or \textit{subject}) and the \textit{tail} (or \textit{object}) entities, and $r\in\mathbb{R}$ is referred to as the \textit{relationship} (or \textit{predicate}).

The problem of KGE is in finding a function that learns the embeddings of triples using low dimensional vectors such that it preserves structural information, $f:\mathbb{D}^+\rightarrow\mathbb{R}^d$. To accomplish this, the general principle is to enforce the learning of entities and relationships to be compatible with the information in $\mathbb{D}^+$. The representation choices include deterministic point~(\cite{bordes2013translating}), multivariate Gaussian distribution~(\cite{He2015}), or complex number~(\cite{Trouillon2016}). Under the Open World Assumption (OWA), a set of unseen \textit{negative} triplets, $\mathbb{D}^-$, are sampled from \textit{positive} triples $\mathbb{D}^+$ by either corrupting the head or tail entity. Then, a  scoring function, $f_r(h, t)$ is defined to reward the positive triples and penalize the negative triples. 
% For example, a loss function normally used in translation based KGE algorithms is as follows: 
% \begin{equation} \label{eq:1}
%     \min_{\mathbb{\theta}} \sum_{(h,r,t)\in\mathbb{D}^+} \sum_{(h',r',t')\in\mathbb{D}^-} \max(0, \gamma - f_r(h, t) + f_{r'}(h',t'))
% \end{equation}
% This loss measures the distance between the translated entities in the embedding space.
Finally, an optimization algorithm is used to minimize or maximize the scoring function. 

KGE methods are often evaluated in terms of their capability of predicting the missing entities in negative triples $(?, r, t)$ or $(h, r, ?)$, or predicting whether an unseen fact is true or not. 
The evaluation metrics include the rank of the answer in the predicted list (mean rank), and the ratio of answers ranked top-k in the list (hit-k ratio).

\section{Software Architecture}
The pykg2vec library is built using Python and TensorFlow. TensorFlow allows the computations to be assigned on both GPU and CPU. In addition to the main model training process, pykg2vec utilizes multi-processing for generating mini-batches and performing an evaluation to reduce the total execution time. The various components of the library (see Figure \ref{fig:archi}) are as follows:

\begin{figure}[h!]
    \centering
     \vspace{-1em}
    \includegraphics[width=0.88\textwidth]{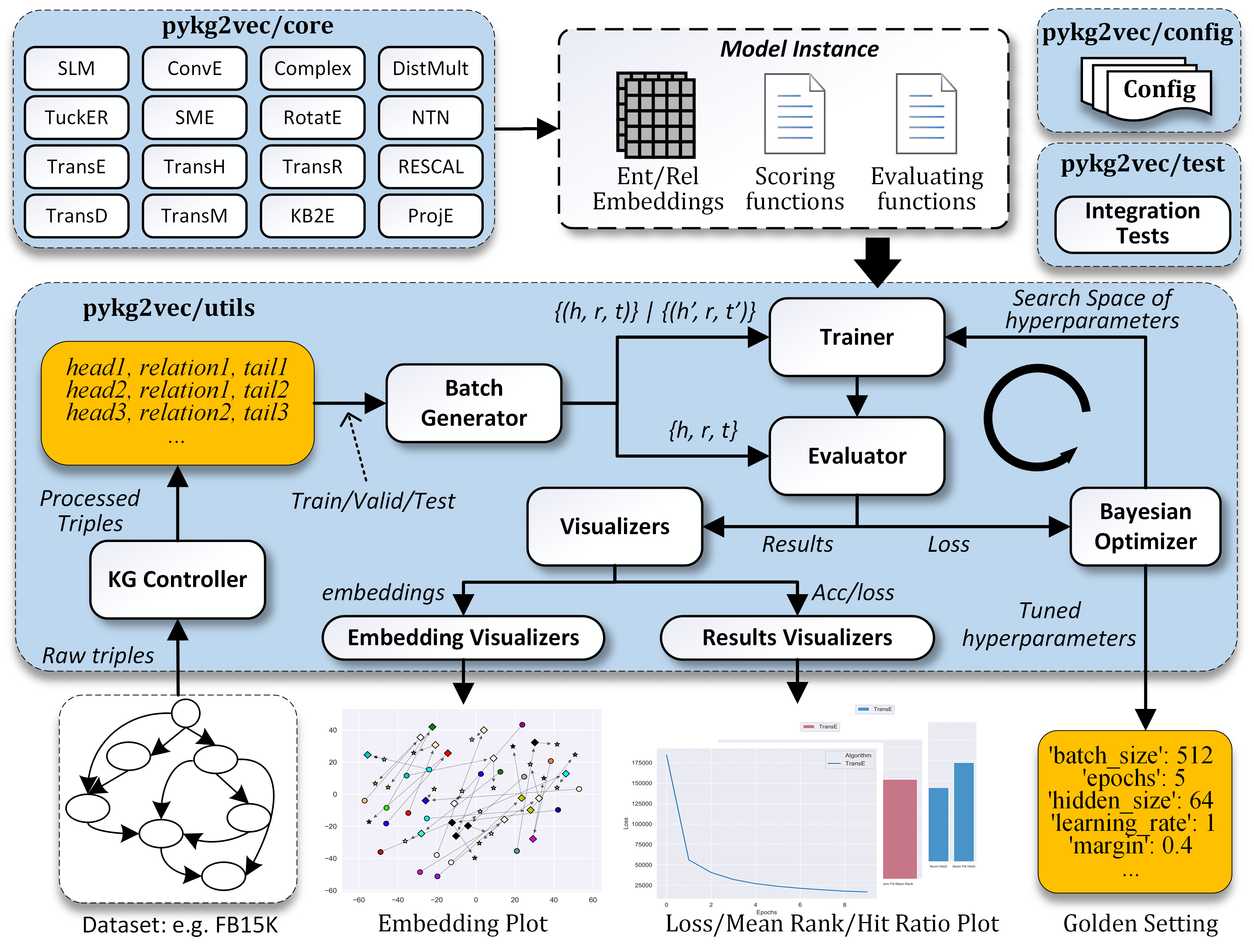}
    \vspace{-1em}
    \caption{Pykg2vec software architecture}
    \vspace{-1.5em}
    \label{fig:archi}
\end{figure}

\begin{itemize}[align=left,leftmargin=*]
\item{KG Controller}: handles all the low-level parsing tasks such as finding the total unique set of entities and relations; creating ordinal encoding maps; generating training, testing and validation triples; and caching the dataset data on disk to optimize tasks that involve repetitive model testing.
% , creating hash map for head and relation pair $(h,r)\rightarrow [t1,t2,t3,\ldots]$ and relation and tail pair $(r,t)\rightarrow [t1,t2,t3,\ldots]$, etc. 
%Then, for each specific dataset, it caches these data on disks so that repetitively testing of the models becomes efficient. 
\item{Batch Generator}: consists of multiple concurrent processes that manipulate and create mini-batches of data. These mini-batches are pushed to a queue to be processed by the models implemented in TensorFlow. The batch generator runs independently so that there is a low latency for feeding the data to the training module running on the GPU.   
% Using a Python module multi-process, pykg2vec separates those tasks from the main training process to separate processes. Once the source of the triplet set has been assigned, the batch generator creates a process and randomly generates a mini-batch and then submits the task via a queue to another client process responsible for generating negative samples.
  
\item{Core Models}: consists of 16 KGE algorithms implemented as Python modules in TensorFlow. Each module consists of a modular description of the inputs, outputs, loss function, and embedding operations. Each model is provided with configuration files that define its hyperparameters.

\item{Configuration}: provides the necessary configuration to parse the datasets and also consists of the baseline hyperparameters for the KGE algorithms as presented in the original research papers.

\item{Trainer and Evaluator}: the Trainer module is responsible for taking an instance of the KGE model, the respective hyperparameter configuration, and input from the batch generator to train the algorithms. The Evaluator module performs link prediction and provides the respective accuracy in terms of mean ranks and filtered mean ranks. 

\item{Visualization}: plots training loss and common metrics used in KGE tasks. To facilitate model analysis, it also visualizes the latent representations of entities and relations on the 2D plane using t-SNE based dimensionality reduction.

\item{Bayesian Optimizer}: pykg2vec uses a Bayesian hyperparameter optimizer to find a golden hyperparameter set. This feature is more efficient than brute-force based approaches.

\end{itemize}

% On top of the mentioned pipeline, interested researchers may then choose to improve the existing models or apply those models on their datasets. For improving the model, the authors might experiment with the methods using separate libraries regardless of the subtle difference between libraries. Or, they may choose to implement a subset of related KGE methods by their own which might spend more time than expected. 

\section{Usage Examples}
\begin{figure}[h!]
\centering
\vspace{-1em}
\begin{tabular}{cc}

  \includegraphics[width=0.40\textwidth]{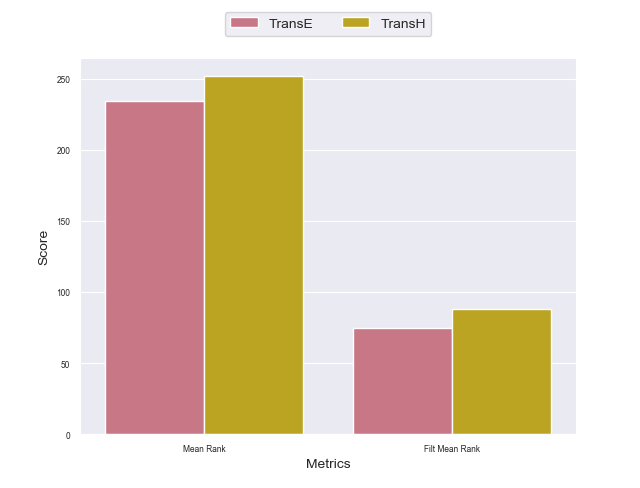}  & \includegraphics[width=0.40\textwidth]{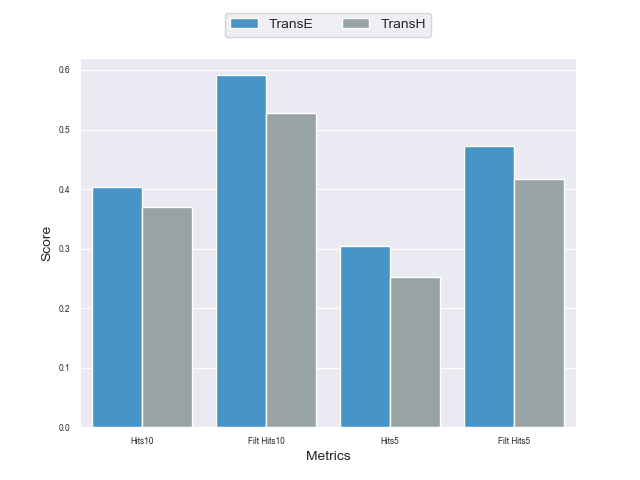}  \\
  (a) Mean ranks of the algorithms & (b) Hit ratios of the algorithms \\
  \includegraphics[width=0.34\textwidth]{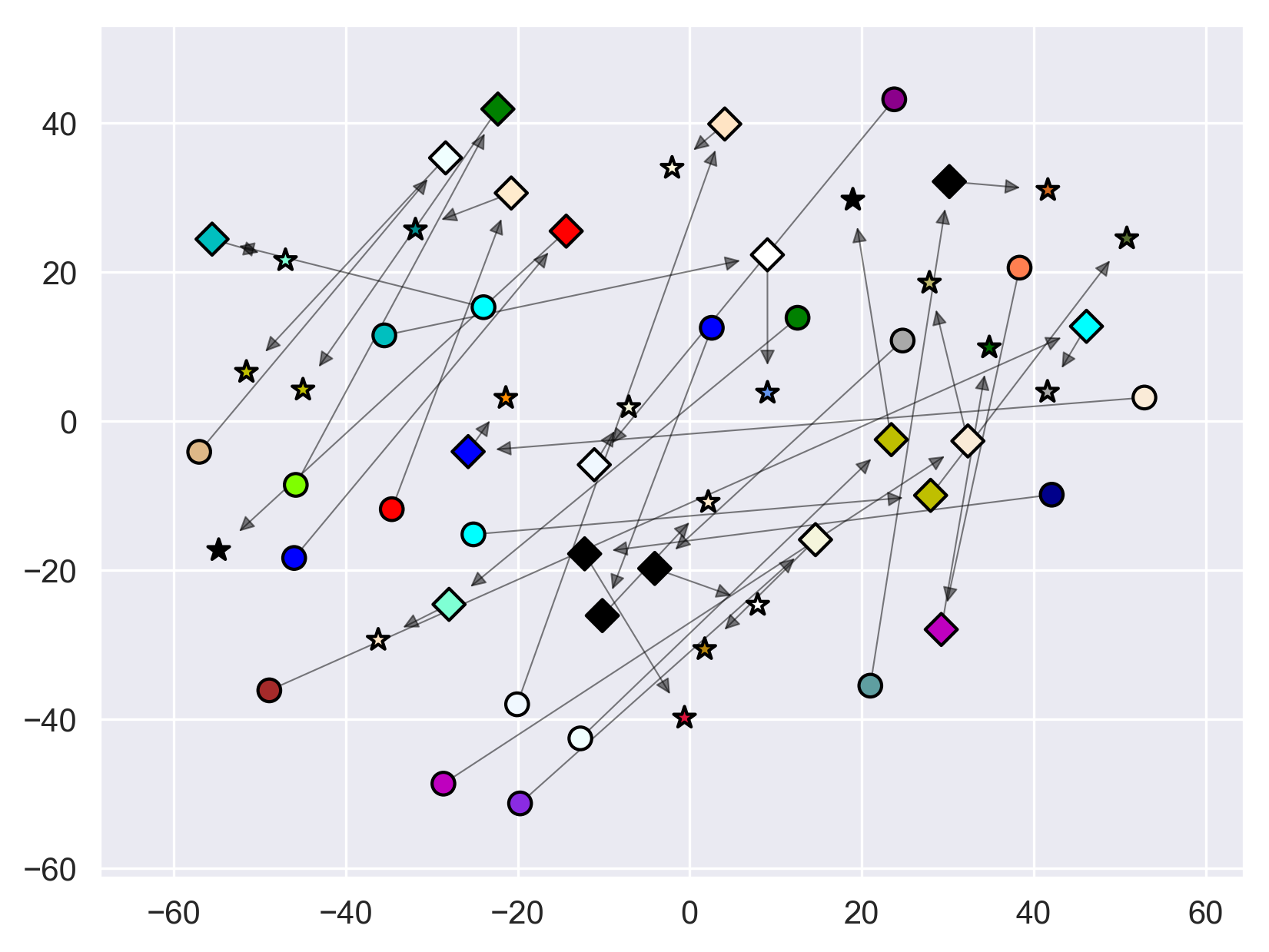}  & \includegraphics[width=0.34\textwidth]{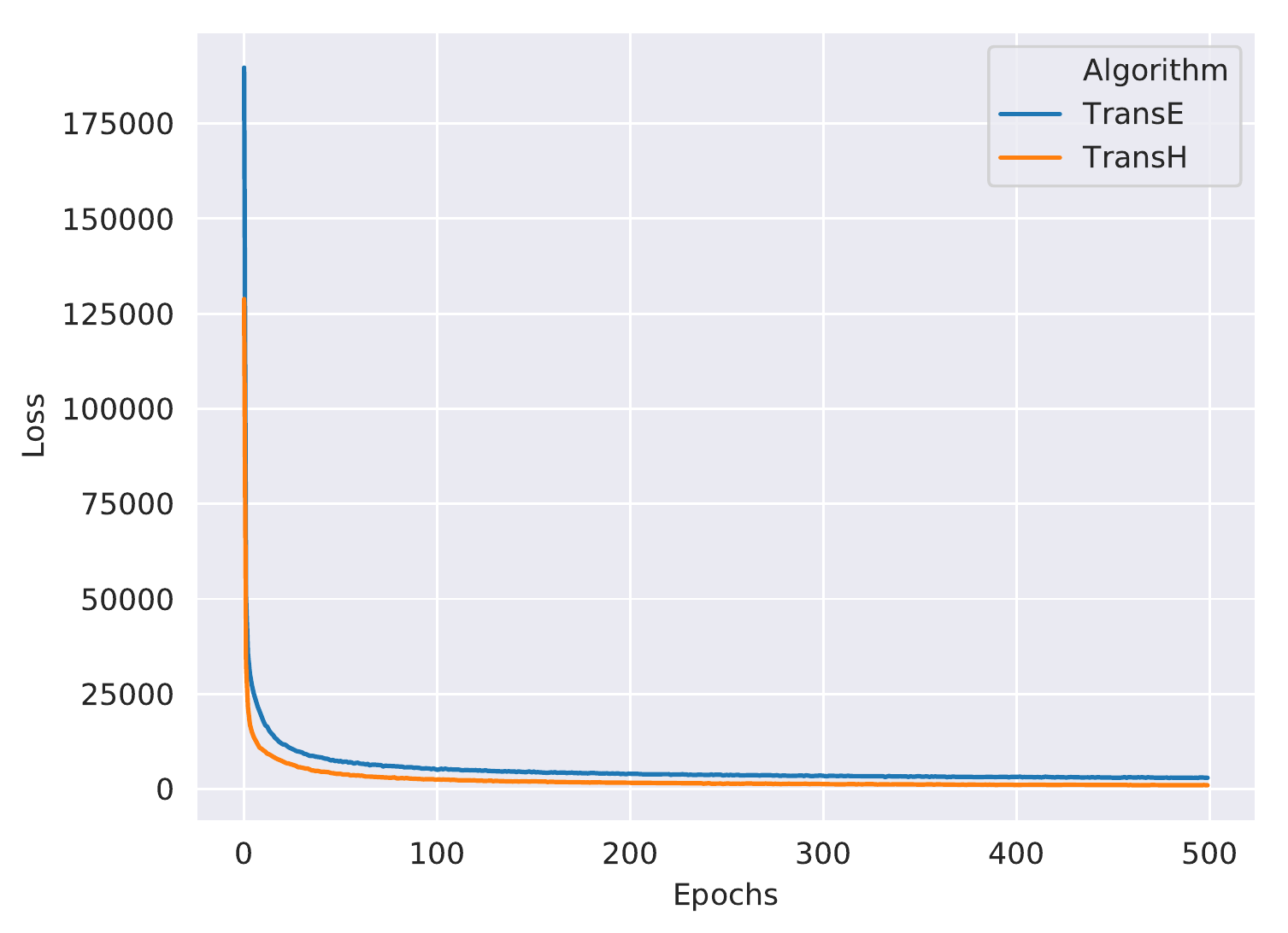} \\ 
   (c) Entity and relation embedding plot & (d) Loss value plot \\
\end{tabular}
\vspace{-0.5em}
\caption{Visualization examples from Freebase15K}
\label{fig:result}
\vspace{-0.5em}
\end{figure}

Pykg2vec provides users with two utilization examples ($train.py$ and $tune\_model.py$) available in the pykg2vec/example folder. Training is performed with the following script:
\begin{python}
$ python train.py -h # Check the manual for input arguments
$ python train.py -mn transe # Train TransE
\end{python}
% $ python train.py -mn transh # Train TransH
To apply the best setting described in the paper, the following script can be invoked.
\begin{python}
$ python train.py -mn transe -ghp True # Train using the golden setting
\end{python}
Some of the results plotted after training $TransE$ and $TransH$  are shown in Figure \ref{fig:result}.

\noindent To tune the model, the following script can be invoked. 
\begin{python}
$ python tune_model.py -mn transe # Tune hyperparameters
Found Golden Setting: # dummy tuning result
{'L1_flag': False, 'batch_size': 256, 'epochs': 5, 'hidden_size': 32,
 'learning_rate': 0.001, 'margin': 0.4, 'opt': 'sgd','samp': 'bern'}
\end{python}
\section{Conclusion}
Pykg2vec is a Python library with extensive documentation that includes the implementations of a variety of state-of-the-art Knowledge Graph Embedding methods and modular building blocks of the embedding pipeline. This library aims to help researchers and developers to quickly test algorithms against their custom knowledge base or utilize the modular blocks to adapt the library for their custom algorithms. 

% Acknowledgements should go at the end, before appendices and references

% \acks{The authors are grateful to the Defense Advanced Research Projects Agency  (DARPA), contract W911NF-17-C-0094, for their support. 
% We would also like to acknowledge support 
% from the National Science Foundation (NSF grant IIS-9988642)
% and the Multidisciplinary Research Program of the Department
% of Defense (MURI N00014-00-1-0637). }


% \bibliography{pykg2vec}

\vskip 0.2in
\begin{thebibliography}{9}
\providecommand{\natexlab}[1]{#1}
\providecommand{\url}[1]{\texttt{#1}}
\expandafter\ifx\csname urlstyle\endcsname\relax
  \providecommand{\doi}[1]{doi: #1}\else
  \providecommand{\doi}{doi: \begingroup \urlstyle{rm}\Url}\fi

\bibitem[Auer et~al.(2007)Auer, Bizer, Kobilarov, et~al.]{Auer2007}
S{\"o}ren Auer, Christian Bizer, Georgi Kobilarov, et~al.
\newblock Dbpedia: A nucleus for a web of open data.
\newblock In \emph{The semantic web}. Springer, 2007.

\bibitem[Bollacker et~al.(2008)Bollacker, Evans, Paritosh, Sturge, and
  Taylor]{Bollacker2008}
Kurt Bollacker, Colin Evans, Praveen Paritosh, Tim Sturge, and Jamie Taylor.
\newblock {Freebase: a collaboratively created graph database for structuring
  human knowledge}.
\newblock \emph{Proc.$\backslash$ of SIGMOD'08}, pages 1247--1250, 2008.
\newblock ISSN 07308078.
\newblock \doi{10.1145/1376616.1376746}.
\newblock URL \url{http://doi.acm.org/10.1145/1376616.1376746}.

\bibitem[Bordes et~al.(2013)Bordes, Usunier, Garcia-Duran, Weston, and
  Yakhnenko]{bordes2013translating}
Antoine Bordes, Nicolas Usunier, Alberto Garcia-Duran, Jason Weston, and Oksana
  Yakhnenko.
\newblock Translating embeddings for modeling multi-relational data.
\newblock In \emph{Advances in neural information processing systems}, pages
  2787--2795, 2013.

\bibitem[Costabello et~al.(2019)Costabello, Pai, Van, McGrath, and
  McCarthy]{ampligraph}
Luca Costabello, Sumit Pai, Chan~Le Van, Rory McGrath, and Nicholas McCarthy.
\newblock {AmpliGraph: a Library for Representation Learning on Knowledge
  Graphs}, March 2019.
\newblock URL \url{https://doi.org/10.5281/zenodo.2595043}.

\bibitem[Free(2014)]{Free2014}
A~Free.
\newblock {Wikidata : A Free Collaborative}.
\newblock pages 1--7, 2014.

\bibitem[Han et~al.(2018)Han, Cao, Lv, Lin, Liu, Sun, and Li]{han2018openke}
Xu~Han, Shulin Cao, Xin Lv, Yankai Lin, Zhiyuan Liu, Maosong Sun, and Juanzi
  Li.
\newblock Openke: An open toolkit for knowledge embedding.
\newblock In \emph{Proceedings of the 2018 Conference on Empirical Methods in
  Natural Language Processing: System Demonstrations}, pages 139--144, 2018.

\bibitem[He et~al.(2015)He, Liu, Ji, and Zhao]{He2015}
Shizhu He, Kang Liu, Guoliang Ji, and Jun Zhao.
\newblock {Learning to Represent Knowledge Graphs with Gaussian Embedding}.
\newblock pages 623--632, 2015.
\newblock \doi{10.1145/2806416.2806502}.

\bibitem[Suchanek et~al.(2017)Suchanek, Kasneci, Weikum, Suchanek, Kasneci,
  Weikum, Core, Suchanek, and Weikum]{Suchanek2017}
Fabian Suchanek, Gjergji Kasneci, Gerhard Weikum, Fabian Suchanek, Gjergji
  Kasneci, Gerhard Weikum, Yago~A Core, Fabian~M Suchanek, and Gerhard Weikum.
\newblock {Yago : A Core of Semantic Knowledge Unifying WordNet and Wikipedia
  To cite this version : YAGO : A Core of Semantic Knowledge Unifying WordNet
  and Wikipedia}.
\newblock 2017.

\bibitem[Trouillon et~al.(2016)Trouillon, Welbl, Riedel, Gaussier, and
  Bouchard]{Trouillon2016}
Th{\'{e}}o Trouillon, Johannes Welbl, Sebastian Riedel, {\'{E}}ric Gaussier,
  and Guillaume Bouchard.
\newblock {Complex Embeddings for Simple Link Prediction}.
\newblock 48, 2016.
\newblock URL \url{http://arxiv.org/abs/1606.06357}.

\end{thebibliography}

\end{document}